%% file: main.tex
\pdfoutput=1
\documentclass[11pt]{article}

\usepackage[final]{acl}
\usepackage{times}
\usepackage{latexsym}
\usepackage{enumitem}
\usepackage{float}

\usepackage[T1]{fontenc}
\usepackage[utf8]{inputenc}
\usepackage{amsmath}
\usepackage{microtype}

\usepackage{inconsolata}

\usepackage{graphicx}
\usepackage{multirow}
\usepackage{booktabs}
\usepackage{natbib}
\usepackage{amsmath}
\usepackage{latexsym}
\usepackage{siunitx}

\usepackage{pifont}

\title{Text Detoxification: Data Efficiency, Semantic Preservation and Model Generalization}

\author{
\textbf{Jing Yu\textsuperscript{1}\thanks{Equal contribution.}},
\textbf{Yibo Zhao\textsuperscript{1}\footnotemark[1]},
\textbf{Jiapeng Zhu\textsuperscript{1}},
\textbf{Wenming Shao\textsuperscript{2}},
\textbf{Bo Pang\textsuperscript{2}},
\textbf{Zhao Zhang\textsuperscript{1}},
\textbf{Xiang Li\textsuperscript{1}\thanks{Corresponding author: \texttt{xiangli@dase.ecnu.edu.cn}}}
\\
\textsuperscript{1}School of Data Science and Engineering, East China Normal University \\
\textsuperscript{2}Shanghai EastWonder Info-tech Co., Ltd. 
}

\begin{document}
\maketitle
\begin{abstract}
The widespread dissemination of toxic content on social media poses a serious threat to both online environments and public discourse, highlighting the urgent need for detoxification methods that effectively remove toxicity while preserving the original semantics.
However, existing approaches often struggle to simultaneously achieve strong detoxification performance, semantic preservation, and robustness to out-of-distribution data. 
Moreover, they typically rely on costly, manually annotated parallel corpora while showing poor data efficiency.
% Moreover, many existing approaches depend heavily on costly, manually annotated parallel corpora, while offering limited data efficiency.
% Moreover, many rely heavily on expensive, manually annotated parallel datasets, with unsatisfied data efficiency.
% which limits scalability.
To address these challenges, we propose a two-stage training framework that jointly optimizes for \textbf{Data Efficiency}, \textbf{Semantic Preservation}, and \textbf{Model Generalization}. 
We first perform supervised fine-tuning on a small set of high-quality, filtered parallel data to establish a strong initialization. 
Then, we leverage \textbf{unlabeled} toxic inputs and a custom-designed reward model to train the LLM using Group Relative Policy Optimization.
Experimental results demonstrate that our method effectively mitigates the trade-offs faced by previous work, achieving state-of-the-art performance with improved generalization and significantly reduced dependence on annotated data. Our code is available at \url{https://github.com/allacnobug/Detoxification-of-Text}.
% Traditional content moderation often misclassifies benign speech, and the practice of blocking or deleting toxic content is criticized for infringing on free expression and distorting public discourse, prompting interest in text detoxification techniques that reduce toxicity while preserving meaning.
% However, existing methods still struggle with detoxification accuracy, semantic consistency, and cross-domain generalization.
% In this paper, we propose a two-stage detoxification framework based on large language models to address these challenges, introducing reinforcement learning into the task of text detoxification for the first time. We first perform a cold start using a small amount of high-quality parallel data. Then, inspired by DeepSeek’s Group Relative Policy Optimization (GRPO), we adopt this reinforcement learning approach with a self-designed reward function that combines semantic similarity and toxicity scores to guide the model toward more effective detoxification during training. This method eliminates the reliance on large-scale parallel corpora while significantly improving the model’s overall performance as well as its adaptability and robustness across cross-domain datasets. Experimental results show that our method achieves the best overall performance across three benchmark datasets, surpassing all existing mainstream approaches.
\end{abstract}

\textcolor{red}{\textbf{Disclaimer}: \textit{This paper describes toxic and discriminatory content that may be disturbing to some readers.}}

\section{Introduction}

\begin{figure}[t]
  \centering\includegraphics[width=0.85\columnwidth]{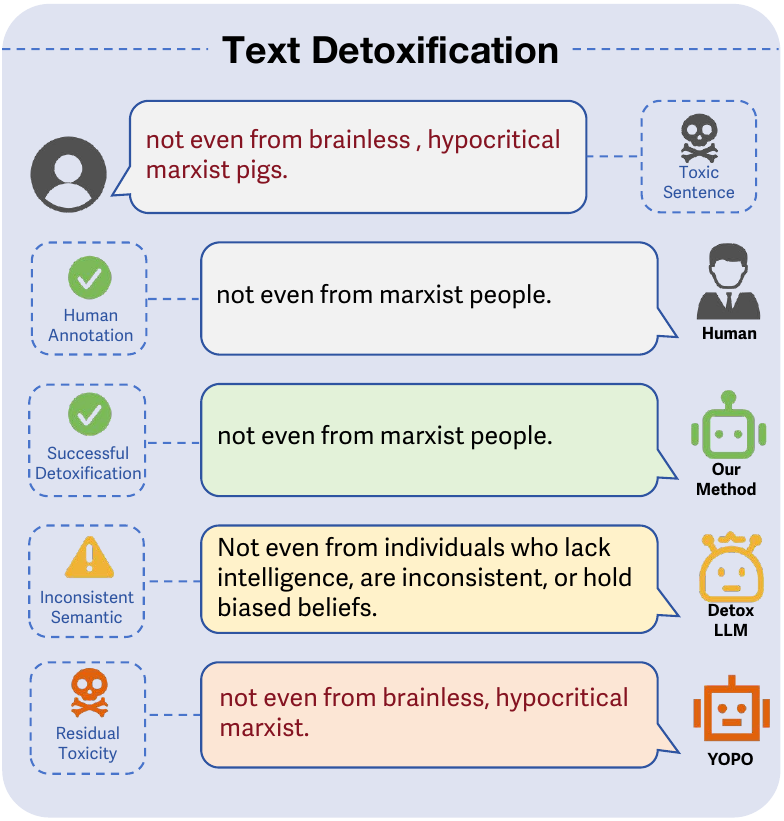}
  \caption{This example demonstrates the current methods fail to balance detoxification and semantic preservation. DetoxLLM and YOPO are representative mainstream approaches, as introduced in Section~\ref{sec:related_work}.}
  \label{fig:datacase}
\end{figure}

% \zyb{\input{intro_new}}

With the rapid growth of online media platforms, an increasing number of users engage in discussions and debates. These interactions often contain toxic content, such as insults, discrimination, or hate speech, which poses significant threats to the digital environment and user experience~\citep{muller2021fanning, muller2023effects, bursztyn2019social,du2023symptom,cao2023can}.
Current moderation mechanisms primarily rely on blocking or removing such content, which can lead to false positives and is widely criticized for infringing on freedom of expression and distorting public discourse~\citep{tworek2021history, habibi2024content}. 
As a result, a key research challenge has emerged: \textit{how to automatically rewrite toxic content into harmless language while preserving the original stance and intent?}

Previous work on this problem has primarily employed relatively small models such as T5~\cite{laugier2021civil,dale2021text,he2024you} and BART~\cite{logacheva2022paradetox}. However, toxic content often involves complex semantics and exhibits highly variable styles, leading to frequent out-of-distribution~(OOD) scenarios. 
Due to their limited capacity for semantic understanding and generalization, these smaller models struggle to produce high-quality rewrites, resulting in suboptimal performance.
Moreover, as illustrated in Fig.~\ref{fig:datacase}, previous approaches often fail to strike a balance between detoxification and semantic preservation. For example, DetoxLLM~\cite{khondaker2024detoxllm} achieves effective detoxification but retains little of the original meaning, undermining the rewritten text's utility in maintaining meaningful community discourse. In contrast, YOPO~\cite{he2024you} largely preserves the original semantics, but its detoxification accuracy is relatively low, allowing toxic content to persist in the community.

The rapid progress of large language models~(LLMs) in recent years offers new insights into this problem.
Thanks to their strong semantic understanding and generalization capabilities, LLMs are particularly well-suited for the task of toxic content rewriting.
Motivated by this, we explore leveraging LLMs to effectively transform toxic inputs into harmless yet semantically equivalent outputs.
Yet, due to the alignment process with human values during the post-training phase, LLMs tend to be highly sensitive to toxic content~\cite{zhang-etal-2024-dont-go, zhao2024metatox}. As a result, naive approaches based on prompt engineering or few-shots often lead to a refusal to generate outputs. Even when outputs are generated, LLMs frequently restructure the entire input, which may cause unnecessary alterations and loss of original meaning. 
To overcome this limitation, it is necessary to fine-tune the model specifically for the detoxification task.

However, manually annotating toxic content is both costly and ethically sensitive. Existing public datasets for toxic content rewriting are limited in size and often suffer from inconsistent quality, with frequent deviations from the original stance. Directly applying supervised fine-tuning (SFT) on such noisy data may lead to a \textit{garbage in, garbage out} effect. Moreover, since SFT primarily encourages memorization rather than generalization~\cite{chu2025sft}, it may further limit the model’s performance. Inspired by the recent success of reinforcement learning in post-training large language models~\cite{grpo,dapo}, we explore reinforcement learning as a more robust alternative for aligning LLMs with the detoxification objective in an annotation-free manner.
Specifically, we first perform SFT using a small amount of carefully filtered, high-quality data to establish a solid initialization.
Then, we leverage unannotated toxic inputs and train the model using GRPO, guided by a reward function that jointly considers semantic similarity and detoxification quality.
This two-stage training paradigm enables us to achieve performance surpassing existing state-of-the-art (SOTA) methods using only a fraction of the annotated data, and demonstrates strong generalization on out-of-distribution (OOD) benchmarks.
Our main contributions can be summarized as follows:

\begin{itemize}[leftmargin=*, itemsep=0pt, parsep=0pt, topsep=0pt, partopsep=0pt]
    
    \item \textbf{Data-efficient detoxification}: We propose a training framework that achieves the best performance using only 20\% of the annotated data, significantly reducing reliance on costly human annotations.

    \item \textbf{Balancing detoxification performance and semantic preservation}: We are the first to simultaneously optimize for both detoxification effectiveness and semantic consistency, achieving state-of-the-art performance across multiple baseline comparisons. 

    \item \textbf{Strong OOD performance}: We are the first to introduce GRPO-based reinforcement learning into the toxic content rewriting task, improving generalization and robustness to the diverse and evolving nature of toxic language.

\end{itemize}

\section{Related Works}
\label{sec:related_work}
\begin{figure*}[t]
  \centering\includegraphics[width=0.85\linewidth]{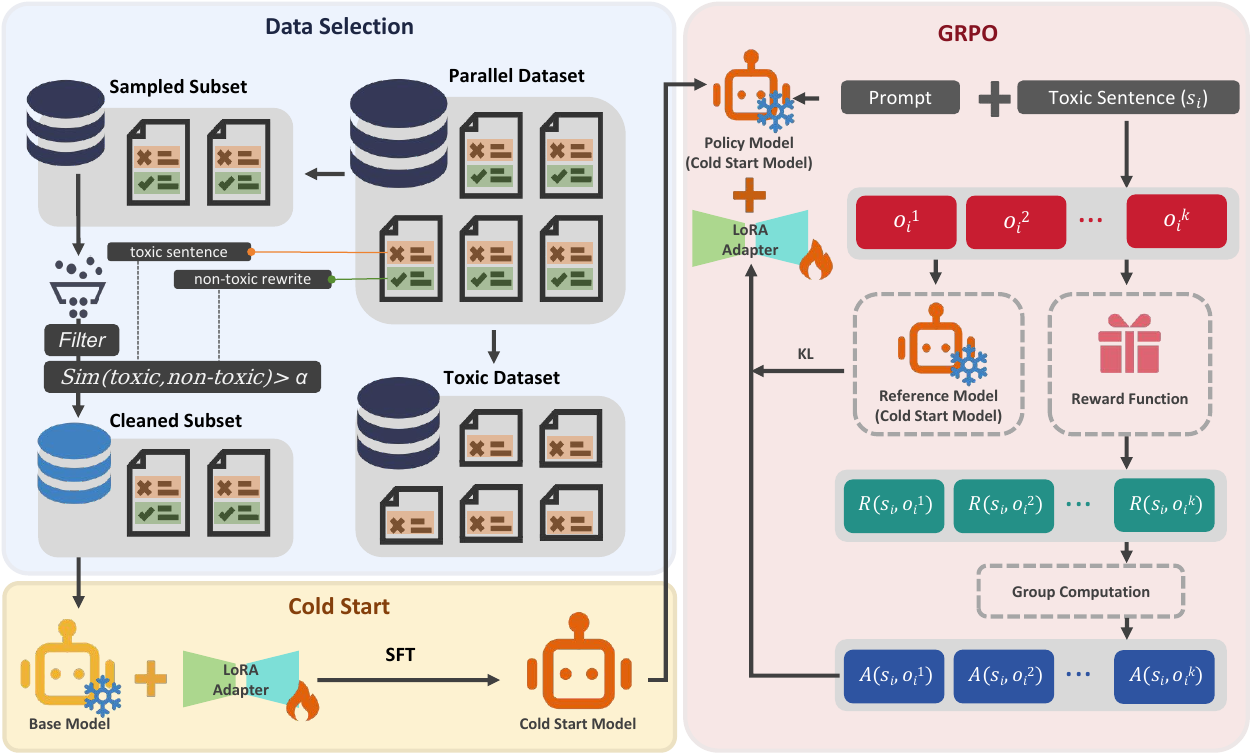}
  \caption{An overview of the model training pipeline, including data selection, cold start by supervised fine-tuning, and reinforcement learning with GRPO.}
  \label{fig:pipeline}
\end{figure*}
\input{related_work_new}

\section{Methodology}

Inspired by the recent success of reinforcement learning in the post-training of LLMs, our approach consists of two main stages: a supervised fine-tuning (SFT) cold start and subsequent annotation-free optimization guided by a designed reward function combining the detoxification quality and the semantic preservation, the whole process is demonstrated in Fig.~\ref{fig:pipeline}. To facilitate a clear presentation of our method, we define a \textit{parallel dataset} as $\mathcal D=\{(s_1,s_1^\prime),\cdots,(s_n,s_n^\prime)\}$, where $s_i$ denotes a toxic input sample and $s_i^\prime$ is its corresponding human-annotated detoxified version.

\subsection{Cold Start}\label{sec:coldstart}

\input{method_coldstart}

\subsection{Annotation-free Optimization}\label{sec:grpo}

\input{method_GRPO}

\section{Experimental Evaluation}

To thoroughly evaluate the effectiveness of our approach, we conduct experiments under both in-domain and out-of-domain settings, and compare our method against a wide range of strong baselines.
% We evaluate the performance of the final model on both in-domain and out-of-domain datasets to assess its overall effectiveness and generalization capability. We further compare our method with a range of existing approaches.
In addition, we conduct ablation studies to examine the contribution of each component.
% , and provide case studies for detailed qualitative analysis of the model's behavior.

\subsection{In-domain Evaluation}

\subsubsection{Experiment Settings}

% {In this section, we present the experimental setup for the in-domain scenario, including the datasets used, the baseline methods for comparison, the evaluation metrics employed, and details about our method setup.

\textbf{Datasets.} We conduct our experiments on the high-quality human-annotated parallel dataset \textit{ParaDetox}, which contains approximately 11k sentence pairs. Following the original split, we use 671 samples for testing.
% and randomly sample 1,000 sentences from the training set as the validation set for hyperparameter tuning. 
% For training, we use 20\% of the annotated training data for SFT-based cold start, and the entire training set (excluding reference rewrites) as unlabeled input for GRPO.
{For SFT-based cold start, we use 20\% of the annotated training data, further filtered to retain only examples with a similarity score greater than $\alpha = 0.5$ (Eq.~\ref{eq:1}). The entire training set (excluding reference rewrites) is used as unlabeled input for GRPO.}

\textbf{Baselines.} We compare our method with three representative existing approaches: 
\textit{ParaDetox}, \textit{DetoxLLM}, and \textit{YOPO}, along with two commonly used sequence-to-sequence models, \textit{T5} and \textit{BART}, fine-tuned on the full training dataset. 
To address the limited use of LLMs in prior work, we additionally include two methods using base LLMs for comparison: five-shot prompting and supervised fine-tuning on the full training set.
Since \textit{ParaDetox} and \textit{DetoxLLM} only release pretrained models without training code, we directly evaluate their released checkpoints. {\textbf{Notably, since DetoxLLM is trained on the DetoxLLM dataset, its evaluation on ParaDetox serves as an out-of-domain test, whereas its performance on the DetoxLLM set in our OOD evaluation constitutes an in-domain test.}}
For \textit{YOPO}, we use the best hyperparameters reported by the authors to train on the full dataset without a validation split. 
\textbf{Evaluation Metrics.}
Following ParaDetox, we use the following evaluation metrics to assess detoxification performance:
\begin{itemize}[leftmargin=*, itemsep=0pt, parsep=0pt, topsep=0pt, partopsep=0pt]
  \item \textbf{Style Accuracy (STA)} measures whether the generated output is classified as “non-toxic” ($1$) or not ($0$), indicating detoxification success. Following prior work, we use a RoBERTa-based classifier trained on the Jigsaw dataset~\citep{logacheva2022paradetox}.
  \item \textbf{Semantic Similarity (SIM)} measures to what extent the generated text preserves the meaning of the original input. It is computed as the cosine similarity between sentence embeddings of the original and detoxified texts, using the model proposed by ~\citet{wieting2019beyond}, which is trained on paraphrase pairs from the ParaNMT corpus to produce high similarity scores for semantically equivalent sentences.
  \item \textbf{Fluency (FL)} measures grammatical acceptability, with $1$ indicating acceptable and $0$ otherwise. It is computed using a RoBERTa-based classifier trained on the CoLA dataset~\citep{warstadt2019neural}, and reported as the proportion of acceptable outputs, aligning with ParaDetox.
  \item \textbf{Joint Score (J)} is defined as the product of STA, SIM, and FL. If either STA or FL is 0---indicating failed detoxification or disfluent output---the J score is 0. The final score is computed by averaging the J scores across all samples, as Eq.~\eqref{eq:score}.
    \begin{equation}
    J = \frac{\sum ( \mathrm{STA} \cdot \mathrm{SIM} \cdot \mathrm{FL} )}{\lvert \mathcal{D}_{\text{test}} \rvert}
    \label{eq:score}
    \end{equation}
\end{itemize}

\textbf{Our Method.} To verify the generality of our method across different backbone models, we conduct training on two widely used instruction-tuned LLMs: Llama3.1-8B-Instruct\footnote{\url{https://huggingface.co/meta-llama/Llama-3.1-8B-Instruct}} and Qwen2.5-7B-Instruct\footnote{\url{https://huggingface.co/Qwen/Qwen2.5-7B-Instruct}}.
For semantic similarity computation in the data selection stage, we use the pre-trained all-miniLM-L6-v2 model\footnote{\url{https://huggingface.co/LeoChiuu/all-MiniLM-L6-v2}}.
For the $\text{NonToxic}$ score in the reward function, we train a BERT-based binary classifier on the training set to predict whether a generated sentence is toxic.
For hyperparameter choice,
we sample 1,000 examples from the training set as a validation set for hyperparameter selection in the SFT stage. In the GRPO stage, we use the full training set without annotations. To ensure fairness, we keep most hyperparameters fixed as default in \texttt{trl} framework and include a sensitivity analysis in Section~\ref{sec:SA}.

\subsection{In-Domain Evaluation}

% Table~\ref{tab:comparison} presents a comparison of model performance on the Paradetox test dataset. 

\begin{table}[ht]
  \centering
  \caption{\label{tab:comparison}
    Evaluation results under the in-domain setting. The best scores are highlighted in \textbf{bold}. \textit{Italicized} SIM and FL scores denote that the metric was computed over valid outputs rather than the entire test set.DetoxLLM, which is followed by $^*$, means it is a ood evaluation.}
    % The number of successfully executed tasks for \texttt{T5}, \texttt{Llama3.1-8B} zero shot and five shot are 630, 393 and 458, respectively.
  \resizebox{\columnwidth}{!}{
    \input{table/eval}

  }
\end{table}

Table~\ref{tab:comparison} presents the evaluation results of our method, various baselines, and human-annotated references.
Notably, both the fine-tuned T5 and Llama3.1-8B-Instruct prompted in a few-shot setting occasionally refuse to generate responses.
Specifically, they refuse to respond on 41 and 213 examples out of 671, respectively.
For fair evaluation, we treat these refusals as failed detoxifications, i.e., assigning an STA of 0. Since such refusals do not yield meaningful outputs, we manually exclude them when calculating FL and SIM.
As Joint Score is defined as the product of STA, SIM, and FL, refusals naturally result in a J score of 0, and thus no additional processing is required.
% Meanwhile, under the few-shot setting, Qwen2.5-7B-Instruct does not refuse to respond, but often generates additional explanatory content despite explicit prompts to avoid it. To ensure an accurate evaluation, we manually remove such extraneous information before computing the corresponding scores.
From the table, we make the following observations:

\begin{table*}[t]
  \centering
  \caption{\label{tab:ood-comparison}
    Evaluation results under the ood setting. The best scores are highlighted in \textbf{bold}. \textit{Italicized} SIM and FL scores denote that the metric was computed over valid outputs rather than the entire test set.
    DetoxLLM, which is followed by $^*$, means it is an in-domain evaluation in DetoxLLM. 
    % The number of successfully executed tasks for the T5 model on the DetoxLLM and HuggingFace datasets is 298 and 467, respectively.
  }

\input{table/ood}
\end{table*}

\textbf{(1) Our method achieves the best overall balance across all metrics.} The models trained with SFT+GRPO achieve the highest Joint Scores (J) of 69.61 and 68.26, outperforming all baselines. 
This indicates that GRPO-based reinforcement learning not only enhances detoxification success (STA), but does so without compromising semantic similarity (SIM) or fluency (FL). Notably, both models surpass the human reference in J, underscoring their overall effectiveness. These results are achieved using only 20\% of the parallel data, highlighting the data efficiency of our approach, and suggesting that RL-based training can serve as a promising solution for detoxification tasks, especially under limited annotation.
% These results demonstrate that reinforcement learning with GRPO not only improves detoxification success (STA) but does so without severely compromising semantic preservation or fluency. The fact that both models surpass the human reference on J (65.36) underscores their overall effectiveness. Notably, these results are achieved using only 20\% of the parallel training data, highlighting the efficiency and generalization potential of our approach.

\textbf{(2) Strong detoxification performance often comes at the expense of semantic fidelity.} 
Several baselines, especially DetoxLLM trained on pseudo-parallel data, achieve high STA scores, demonstrating strong detoxification ability. However, this often comes at the cost of semantic fidelity---reflected in their significantly lower SIM scores---since pseudo-parallel data may not preserve the original meaning. In contrast, our method explicitly incorporates a semantic similarity term in the reward function, guiding the model to maintain the original intent during rewriting. As a result, it achieves both high STA and superior SIM scores, even outperforming human rewrites in semantic preservation.
% Several models, notably DetoxLLM and Qwen2.5-7B-Instruct zero-shot, achieve high style accuracy, 95.38 and 96.43, respectively, indicating that their outputs are consistently classified as non-toxic. However, these gains are undermined by low semantic similarity scores (59.15 and 68.38), suggesting that detoxification is achieved via aggressive rewriting. This trade-off is detrimental {\color{red}in tasks where content preservation is critical}, as reflected in their reduced joint scores. In contrast, our fine-tuned models maintain high STA while preserving over 82\% semantic similarity, {\color{red}highlighting the importance of training methods that explicitly model the tension between safety and meaning.}

\textbf{(3) Fluency is saturated across models and thus less discriminative at the top end.} 
All methods achieve relatively high FL scores, comparable to or exceeding human-level performance, indicating that pretrained language models are inherently capable of producing grammatically well-formed and fluent outputs. These results indicate that fluency alone is not a sufficient indicator of detoxification quality. Notably, our method maintains strong fluency without explicit optimization for this metric during training. This is likely attributed to the clip function and KL divergence constraint in the optimizing phase, which prevents the policy from deviating too far from the reference model and thus preserves its generation quality.

In summary, our method outperforms human annotators across all four metrics on a dataset with high-quality annotations and achieves higher semantic preservation while maintaining a strong detoxification success rate. Although our method shows a slight decrease in FL, it remains comparable to or better than human annotators, which is relatively acceptable.
% Most models achieve high fluency scores (\textsc{FL} > 87), with several exceeding 96 (e.g., DetoxLLM, Qwen2.5-7B variants), suggesting that LLMs inherently generate grammatically well-formed outputs due to diverse pretraining corpora. However, high fluency does not necessarily imply strong detoxification performance: for instance, LLaMA3.1-8B (zero-shot) achieves an \textsc{FL} of 98.98 but records the lowest \textsc{STA} (58.27), while YOPO also shows high fluency (87.03) with relatively weak \textsc{STA} (82.27). These results indicate that fluency alone is not a sufficient indicator of detoxification quality, particularly for large models. Notably, our models also maintain strong fluency despite receiving no explicit supervision on the \textsc{FL} metric during training. 

% \textbf{(4) Zero-/few-shot prompting yields unstable and unreliable outputs.} The performance of LLaMA3.1-8B and Qwen2.5-7B under zero-/few-shot prompting is highly inconsistent. While certain metrics appear strong (e.g., \textsc{STA} for Qwen2.5-7B zero-shot), other outputs exhibit incompleteness or formatting issues, failing to objectively reflect the true capabilities of the models. These deficiencies—particularly prominent in T5 and LLaMA3.1—indicate poor output coverage and raise concerns about reliability in real-world applications. Manual post-processing was required to obtain usable outputs for evaluation, underscoring that zero-shot prompting, without task-specific tuning, remains insufficient for robust and consistent detoxification.

\subsection{Out-of-Domain Evaluation}

\subsubsection{Experiment Settings}

Compared to the in-domain setting, the out-of-domain (OOD) experiments differ only in the choice of test data. We directly evaluate the models trained with in-domain settings on two additional datasets: DetoxLLM\footnote{\url{https://huggingface.co/UBC-NLP/DetoxLLM-7B}} and the HuggingFace text detoxification dataset\footnote{\url{https://huggingface.co/datasets/narySt/text_detoxification_dataset}}. Due to the relatively low annotation quality of these datasets, we do not use them for training. Instead, we randomly sample from them to construct OOD test sets. \textbf{Notably, as mentioned in in-domain settings, DetoxLLM on the DetoxLLM dataset is an in-domain test.}

\subsubsection{Out-of-Domain Evaluation}

Table~\ref{tab:ood-comparison} reports model performance on the OOD test sets. We summarize three key findings:
% —\textit{DetoxLLM} and \textit{HuggingFace}—which are used to assess generalization capability.

\textbf{(1) Surpassing the quality of original dataset annotations.} Our method achieves a higher J score compared to the original references. On the DetoxLLM dataset, this improvement is primarily attributed to our method's ability to preserve semantic similarity better. In contrast, on the HuggingFace dataset, the improvement mainly stems from a higher detoxification success rate. These results highlight the potential of our method to serve as a viable alternative to human annotation in future dataset construction.

\textbf{(2) Consistently balancing detoxification quality and semantic preservation.} Even on OOD data, our method strikes a balance between detoxification quality and semantic fidelity, outperforming previously proposed approaches. This enables better retention of the original communicative intent and supports healthier discourse with the community.
Notably, our method surpasses DetoxLLM even in OOD settings, despite DetoxLLM being evaluated in-domain, demonstrating our approach's strong generalization and detoxification capability.

\textbf{(3) Generalization primarily stems from GRPO.} Compared to standard supervised fine-tuning on the full dataset, our method demonstrates superior generalization performance. When using the same Qwen backbone, our method achieves J score improvements of 11.37 and 5.79 over SFT on the DetoxLLM and HuggingFace datasets, respectively.
These results reinforce previous findings: \textit{SFT memorizes, RL generalizes}~\cite{chu2025sft}.

In summary, our method notably outperforms baselines in OOD settings, demonstrating stronger adaptability to the dynamic nature of toxic content in the real world and greater practical utility.

\subsection{Ablation Study}
\begin{table}[t]
  \centering
  \caption{\label{tab:ablation-study}
    Ablation study results based on Llama3.1-8B-Instruct, ParaDetox dataset. 
    % All methods are applied on top of the Llama3.1-8B-Instruct base model. 
    The best scores are highlighted in bold. Italicized SIM
    and FL scores denote that the metric was computed over
    valid outputs.
    % rather than the entire test set.
    % Notably, the Zero-Shot and GRPO methods failed to complete certain tasks or generate outputs in some cases. Consequently, except for STA (based on manually filtered results), other metrics(italicized text) such as SIM, FL, and J were calculated only on successful outputs. These metrics thus reflect performance on the evaluable subset rather than the full task set, and should be interpreted alongside execution success rates.
  }
  \resizebox{\columnwidth}{!}{
    \input{table/ablation}
  }
  
\end{table}

{To understand the contributions of each component, we conduct an ablation study on {Llama3.1-8B-Instruct} model by progressively removing individual modules from our pipeline, the results are illustrated in Table~\ref{tab:ablation-study}.
{Notably, both the zero-shot prompted Llama3.1-8B-Instruct model and the same model trained via GRPO occasionally refuse to generate responses, failing to respond on 278 and 99 out of 671 examples, respectively.}

When the Data Selection module is removed, we observe a notable drop in semantic similarity and other evaluation metrics. This validates our earlier hypothesis: rewrites with low semantic similarity are unlikely to be high-quality rewrites, highlighting the importance of filtering for semantic consistency.
When we further remove the SFT stage, we observe an increase in the FL metric, but a substantial drop in the other two metrics. This indicates that GRPO alone, without the foundational training provided by SFT, fails to effectively solve the task. It highlights the strong dependence of RL training on the base model’s prior capabilities. When the task is out of the capabilities of the base model, the benefits of RL become limited---echoing findings in recent literature questioning the standalone efficacy of RL in such scenarios~\cite{RL1,RL2, RL3}.}

\subsection{Parameter Sensitivity Analysis}
\label{sec:SA}
\input{table/dataproportion}

% \begin{table*}[ht]
%   \centering
%   \resizebox{\textwidth}{!}{%
%   \input{table/casestudy}
%   }
%   \caption{Case study comparison between models on two representative toxic inputs. \ding{168} indicate the baseline model. Toxic terms are highlighted in bold, while italicized text denotes instances where the model exhibits tendencies of over-editing.}
%   \label{tab:case-study}
% \end{table*}

\section{Conclusion}

% In this papar, We aim to address four core challenges in the task of text detoxification: reliance on large-scale parallel corpora, insufficient detoxification accuracy, limited cross-domain generalization, and inadequate semantic preservation. To this end, we propose a two-stage text detoxification framework based on LLaMA3.1-8B, which integrates small-scale cold-start fine-tuning with a GRPO-based reinforcement learning strategy. Our approach achieves dual optimization of detoxification accuracy and semantic consistency without relying on large-scale parallel data.
% Experimental results demonstrate that the proposed method significantly outperforms existing mainstream models across multiple benchmark datasets. It not only outperforms other mainstream methods in in-domain settings but also demonstrates strong generalization and robustness in out-of-domain scenarios.

In this paper, we identified three major limitations of current detoxification approaches: heavy reliance on manually annotated parallel corpora, inability to balance detoxification quality and semantic preservation, and limited generalization capability. To address these challenges, we proposed a reinforcement learning approach that simultaneously optimizes detoxification effectiveness and semantic preservation, without requiring large-scale annotated data. Experimental results show that our method effectively overcomes the aforementioned issues and even surpasses human-annotated references across multiple benchmarks.
% Our findings shed light on the potential of leveraging our framework---combined with techniques such as rejection sampling---for automated data labeling, paving the way for further improvements in model performance.

\section*{Limitations}

Despite the promising results, our approach still has several limitations. 
(1) \textbf{Generalization to noisy out-of-domain data remains limited}: While our model shows improved generalization compared to baseline methods, as demonstrated in Table~\ref{tab:ood-comparison}, it still struggles with out-of-domain inputs containing noisy elements such as URLs, usernames, or emojis, which are present in the DetoxLLM dataset. These complex and irregular patterns pose challenges for effective detoxification. 
(2) \textbf{Handling of implicit toxicity is weak}: This is primarily due to the lack of implicit toxic examples in the training dataset. Furthermore, implicit toxicity is inherently difficult to detect and neutralize, as its harmful meaning is often embedded within subtle semantics—sometimes even beyond human annotators’ judgment. 
(3) \textbf{Fluency is slightly sacrificed for semantic preservation}: As shown in Table~\ref{tab:comparison}, the fluency of outputs generated by the fine-tuned model is slightly lower than that of the base model. This suggests a trade-off where preserving meaning during detoxification may come at the expense of output naturalness. 
% (4) \textbf{Current evaluation metrics are insufficient}: Although we employ four metrics—STA, SIM, FL, and J—that reflect different aspects of detoxification, they do not fully capture all relevant dimensions. A more comprehensive and fine-grained evaluation framework is needed for more accurate assessment of detoxification quality.

\section*{Ethics Statement}

This study aims to perform non-toxic rewriting (detoxification) of toxic online texts while preserving their original semantics as much as possible, in order to reduce the negative impact of toxic language on the online environment and broader social discourse. We hope that by enhancing the safety of content generated by language models, our work can contribute to building a healthier and more inclusive space for online communication.

We acknowledge that the definition of “toxicity” is inherently subjective and context-dependent, with varying standards across different cultural and linguistic backgrounds. To mitigate these challenges, we employed publicly available and structurally standardized datasets to ensure the clarity and consistency of our task objectives. All training and evaluation data used in this study are anonymized and drawn from public sources, containing no personally identifiable or sensitive information. We did not train on any private or unauthorized datasets.

We are also aware of the potential dual-use risks of detoxification systems. Such models could be misused to obscure harmful intent and evade content moderation mechanisms. To prevent this, we recommend that detoxification systems be deployed in conjunction with human oversight, and we emphasize the importance of transparency and accountability in their application.

Finally, we recognize that our system still faces limitations in handling complex semantics, implicit toxicity, and multilingual inputs. We welcome further research and critical evaluation from the community to improve this method and to contribute to the responsible development of AI technologies.

The datasets we used are all existing open-source datasets, aligning with their intention for scientific research. We also adhered to the OpenRAIL++ license for the ParaDetox dataset, followed the MIT license for the Hugging Face Text Detoxification Dataset, and used the DetoxLLM dataset for academic research purposes only, as its license is not explicitly specified.

% Bibliography entries for the entire Anthology, followed by custom entries
%\bibliography{anthology,custom}
% Custom bibliography entries only
\bibliography{custom} 

\newpage
\appendix
\section{Prompt and Parameters}
\label{sec:pp}

\subsection{Prompt}

\begin{figure}[H]
    \centering
    \includegraphics[width=\textwidth]{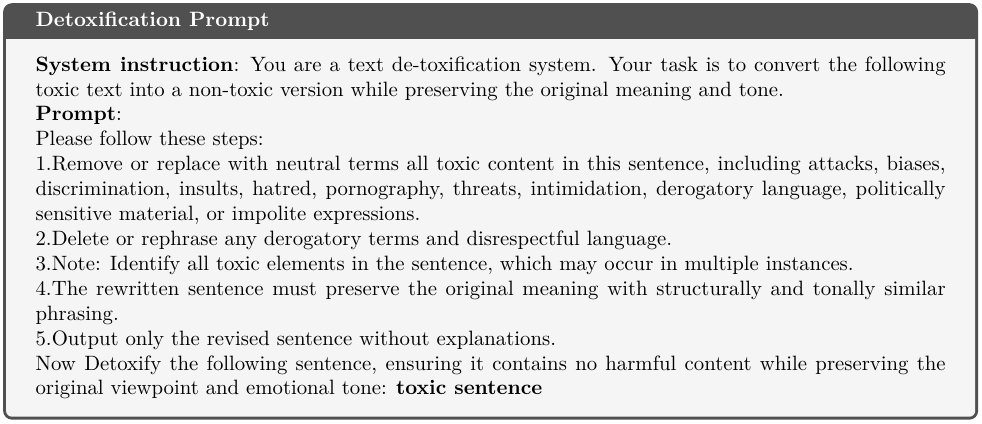} 
    \label{fig:prompt}
\end{figure}
\begin{figure}[H]
    \centering
    \includegraphics[width=\textwidth]{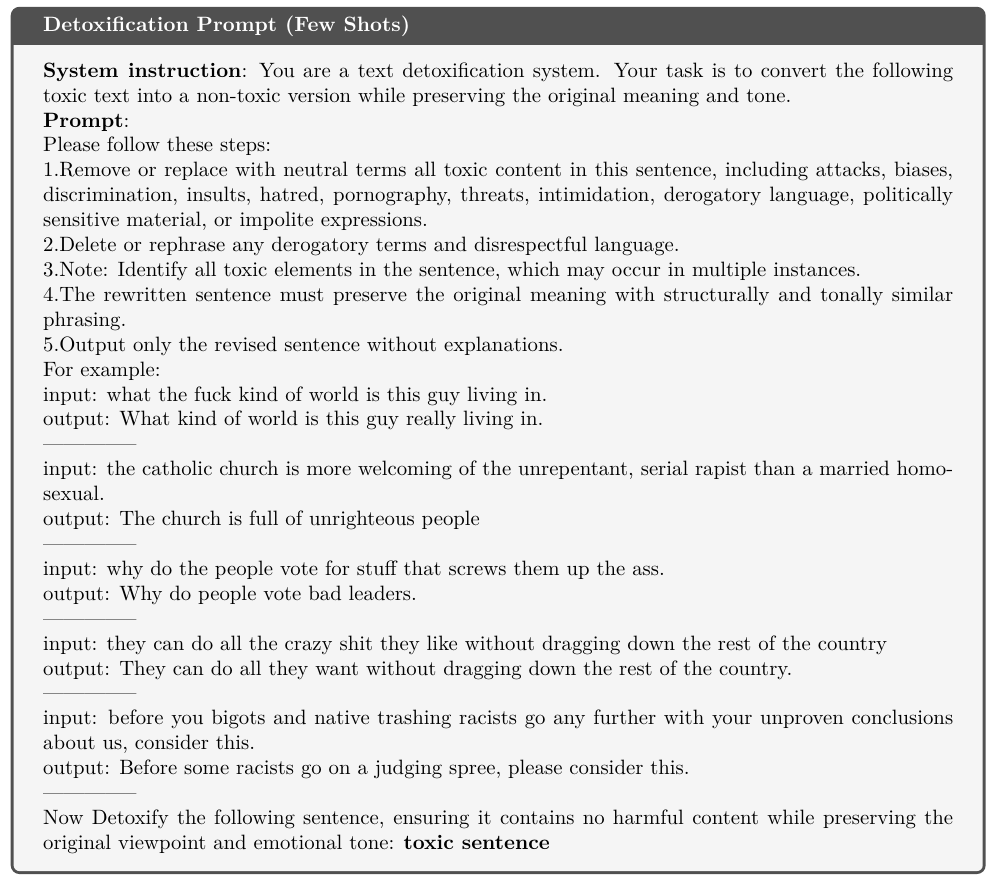} 
    \label{fig:prompt_fs}
\end{figure}

\clearpage

\subsection{Parameters in Cold Start}

During the cold start, we apply LoRA training to the Llama3.1-8B-Instruct model and the Qwen2.5-7B-Instruct model. The optimizer is Adam with an initial learning rate of 5e-5, scheduled using cosine decay. A warm-up of 20 steps is applied to stabilize early training. Gradient accumulation steps are set to 8, with a total of 3 training epochs. All other hyperparameters follow the default settings.

\subsection{Parameters in GRPO}

In the GRPO stage, we apply LoRA training to the models after cold-start. The learning rate is scheduled with cosine decay, starting at 1e-5, with 10\% of total steps used for warm-up. L2 regularization is applied with a weight decay of 0.1. We use a gradient accumulation step of 8 and a maximum gradient norm of 1. During generation, four candidate outputs are generated per input and the temperature in sampling is set to 2.0. The number of training epochs is set to 5 and other parameters remain at their default values. All experiments were conducted using a single NVIDIA A800 GPU.

% \section{Data Select}
% \label{sec:data_select}
% \input{table/dataselect}

\section{Parameter Sensitivity Analysis}
\label{sec:PSA}
\input{table/atest}
\input{table/alpha}

\end{document}

%% file: related_work_new.tex
% \subsection{Toxic Content Detoxification}
Existing detoxification methods can be broadly categorized into two groups: \textbf{unsupervised} and \textbf{supervised} approaches.

Unsupervised methods rely on \textbf{non-parallel datasets}. Such datasets are relatively easy to collect, as they do not require one-to-one semantic alignment.
Representative works in this category include ParaGeDi~\cite{dale2021text}, which combines a T5-based paraphraser for semantic preservation with a GPT-2 discriminator to guide the generation of non-toxic outputs. Another example is CAE-T5~\cite{laugier2021civil}, which frames detoxification as a denoising auto-encoding task, eliminating the need for parallel supervision altogether.
However, these methods typically suffer from poor semantic preservation, often generating outputs that deviate significantly from the original intent, making them impractical for real-world use.

% Since unsupervised methods often struggle with semantic preservation, most researchers leverage machine translation tools~\citep{briakou2021xformal} and pretrained language models~\citep{zhang2020parallel} to perform style transfer when parallel data is available. Although parallel datasets are costly to obtain, they have been shown to significantly enhance detoxification performance. 
Supervised methods, on the other hand, depend on \textbf{parallel datasets}, where each toxic sentence is paired with a manually rewritten non-toxic version that maintains the original meaning. While parallel datasets can improve semantic preservation, their construction is prohibitively expensive, and even manual rewriting does not ensure high-quality data.
A representative work in this category is ParaDetox~\cite{logacheva2022paradetox}, which constructs a small-scale parallel corpus and trains a model on it, outperforming unsupervised methods in both fluency and faithfulness. Building on this, YOPO~\cite{he2024you} applies prompt tuning on the same dataset to guide a T5 model for detoxification.
However, both methods are trained on limited annotated data, resulting in insufficient generalization capabilities and suboptimal performance.

To address data scarcity, DetoxLLM~\cite{khondaker2024detoxllm} leverages ChatGPT to generate large-scale pseudo-parallel data, enabling the training of a 7B-parameter LLM and achieving state-of-the-art results at the time. However, the pseudo labels often fail to preserve semantic fidelity, as ChatGPT tends to detoxify by rewriting entire sentences, which in turn leads to suboptimal semantic preservation in the final model. Moreover, DetoxLLM is costly and still struggles with generalization, particularly under OOD settings, limiting its applicability in diverse real-world scenarios.

In contrast to existing approaches, our method is the first to introduce reinforcement learning into the toxic content detoxification task and achieves data efficiency.
Our method requires only 20\% of manually annotated parallel data for supervised training.
By leveraging a reward function that jointly considers detoxification quality and semantic preservation, our approach significantly reduces the reliance on expensive parallel datasets while maintaining strong detoxification performance and minimizing semantic drift from the original input.

%% file: method_coldstart.tex
%1. 冷启动

To help the model better follow task instructions and gain a preliminary understanding of our detoxification objective, we first perform supervised fine-tuning. However, to avoid overfitting the model to exact input-output mappings---which could reduce diversity during exploration and hinder reinforcement learning---we randomly sample a small subset of the parallel dataset for SFT rather than using the full data, denoted as $\mathcal D_\text{sampled}$.

Given the limited amount of data used in the cold-start stage, data quality becomes particularly crucial. Therefore, we apply a data filtering process to the sampled subset to ensure the reliability and effectiveness of the supervision signal. Specifically, some samples in the training data exhibit significant semantic drift between the original toxic input and its detoxified counterpart.
To mitigate this, we introduce a semantic similarity threshold $\alpha$. 
We first leverage a pre-trained Sentence-BERT model to encode both $s_i$ and $s_i^\prime$, and then calculate the similarity score.
% based on their embeddings.
Pairs with similarity scores above $\alpha$ are retained, while those below are discarded. This filtering step ensures that the supervision signal in the SFT stage maintains semantic consistency, providing a reliable foundation for subsequent reinforcement learning. Formally, the above filtering process is defined as Eq.~\eqref{eq:1}.
\begin{equation}
\mathcal D_{\text{filtered}} = \{(s_i, s_i') \in \mathcal D_\text{sampled} \mid \text{sim}(s_i, s_i') \geq \alpha \}
\label{eq:1}
\end{equation}
After the data filtering step, we use the cleaned dataset $\mathcal D_\text{filtered}$ for instruction-tuned supervised training. Specifically, for each pair $(s_i,s_i^\prime) \in \mathcal D_\text{filtered}$, we concatenate an instruction prompt with the toxic input $s_i$ as the model input, and use the corresponding detoxified sentence $s_i^\prime$ as the target output. We fine-tune the model using LoRA to efficiently adapt the base model parameters with minimal computational overhead. Details of the instruction prompt template and LoRA hyperparameters can be found in App.~\ref{sec:pp}.

%% file: method_GRPO.tex
To enhance the model's generalization ability
, we perform post-training using GRPO, an online reinforcement learning method that iteratively improves the model using its own generated data during training, guided by a reward function that jointly captures both semantic similarity and detoxification quality. Before detailing GRPO, we first introduce the design of the reward function, which plays a central role in steering the learning process.

In our task, the ideal output should satisfy two key criteria:~(1) it must be detoxified, i.e., \textit{strong detoxification quality}, and (2) it must preserve the semantic intent and stance of the original sentence, i.e. \textit{faithful semantic preservation}. To capture both aspects, we define a composite reward function.

For \textit{detoxification quality}, we train a BERT-based classifier on a labeled dataset to distinguish toxic and non-toxic sentences. The classifier outputs the probability that a given sentence is non-toxic, which we denote as $\text{NonToxic}(\cdot)$. For \textit{semantic preservation}, we employ a pre-trained Sentence-BERT model to compute the semantic similarity between the generated output and the original sentence, denoted as $\text{Sim}(\cdot, \cdot)$. We then define the final reward for a generated output $o_i$, conditioned on the original toxic input $s_i$, as a weighted combination of the two components:
\begin{equation}
    R(s_i,o_i) = \lambda\cdot \text{NonToxic}(o_i)+\text{Sim}(s_i,o_i)
\label{eq:reward}
\end{equation}
Here, $\lambda$ is a hyperparameter that balances the importance of \textit{detoxification quality} and \textit{semantic preservation}.

Following the reward guidance, we proceed to train the model using GRPO. Specifically, our method consists of four key steps.

\textbf{Generation Completions}: For each toxic sentence $s_i$ in the training set, we first use a unified prompt template to instruct the model---initialized via SFT and optimized during the GRPO stage\textemdash to generate $k$ candidate outputs. This results in a set of pairs $\{(s_i,o_i^1),(s_i,o_i^2),\cdots,(s_i,o_i^k)\}$, where each $o_i^j$ represents a potential detoxified version of $s_i$. The details of the unified prompt template can be found at App.~\ref{sec:pp}.

\textbf{Advantage Computation}: To stabilize training, we introduce a baseline strategy. Specifically, for each toxic input $s_i$ with $k$ generated outputs $\{o_i^1,o_i^2,\cdots,o_i^k\}$, we compute the corresponding reward values $\{R(s_i,o_i^1),\cdots, R(s_i,o_i^k)\}$, denoted as the reward distribution $R_{s_i}$. We then normalize each reward by subtracting the mean and dividing it by the standard deviation of this distribution. This normalization reduces variance in the reward signal, providing a more stable training signal and encouraging the model to generate outputs that outperform the average candidate for a given input. The advantage function $A(\cdot,\cdot)$ can be formally defined as Eq.~\eqref{eq:adv}, where $\mu(\cdot)$ and $\sigma(\cdot)$ denote the mean and standard deviation, respectively.
\begin{equation}
    A(s_i,o_i^j) = \frac{R(s_i,o_i^j) - \mu(R_{s_i})}{\sigma(R_{s_i})}
\label{eq:adv}
\end{equation}
Following the GRPO, we assign the same advantage score to all tokens within a sequence, i.e., $A_{i,t}^j = A(s_i,o_i^j), \quad \forall t\in[1,2,\cdots,\text{len}(o_i^j)]$. 
% This uniform assignment ensures consistency between the sequence-level reward and the token-level optimization.

\textbf{KL Divergence Estimation}: To prevent the model from drifting too far from its initial distribution and collapsing during training, we incorporate a token-level KL divergence penalty. Following the GRPO framework, we adopt the k3 estimator as a surrogate for the KL divergence term. The k3 loss is particularly suitable for this setting, as it is both unbiased and exhibits low variance, making it an effective and stable choice for regularization during policy optimization. 
Specifically, $\pi_\theta(o_{i,t}^j|p, o_{i<t}^j)$ and $\pi_\text{ref}(o_{i,t}^j|p, o_{i<t}^j)$ denote the probabilities assigned to token $o_{i,t}^j$ by the current model and the reference model, respectively, given the prompt $p$ and the previously generated token $o_{i<t}^j$. Here, the reference model refers to the model checkpoint obtained after the cold-start SFT phase, serving as a stable baseline to regularize the learning process. The token-level KL divergence can be formally described as Eq.~\eqref{eq:kld}.
    \begin{equation}
    \label{eq:kld}
    \begin{split}
    D_{\mathrm{KL}}(\pi_\theta \| \pi_{\mathrm{ref}})_{i,t}^j = r_{i,t}^j - 1 - \log r_{i,t}^j
    \end{split}
    \end{equation}
where the policy ratio $r_{i,t}^j$ is Eq.~\eqref{eq:ratio}:
\begin{equation}
r_{i,t}^j = \frac{\pi_\theta(o_{i,t}^j \mid p, o_{i<t}^j)}{\pi_\text{ref}(o_{i,t}^j \mid p, o_{i<t}^j)}\label{eq:ratio}
\end{equation}

\textbf{Loss Calculation}: The objective of GRPO is to maximize the advantage while constraining the model to remain close to the reference policy, thereby ensuring training stability. 
Following PPO and GRPO, we adopt the \textbf{clipped surrogate objective} to ensure stable updates and prevent the policy from deviating excessively from the reference model. The objective for each token is defined as Eq.~\eqref{eq:litj}:
\begin{equation}
l_{i,t}^j = \min \left(
    r_{i,t}^j  A_{i,t}^j,\,
    \text{clip}(r_{i,t}^j, 1 - \epsilon, 1 + \epsilon) A_{i,t}^j
\right)\label{eq:litj}
\end{equation}
where the $r_{i,t}^j$ is the policy ratio defined in Eq.~\eqref{eq:ratio}.

The clipping function $\text{clip}(r, 1-\epsilon, 1+\epsilon)$ bounds the policy ratio within a safe range, ensuring that updates are conservative when the new policy significantly diverges from the reference policy. This mechanism helps to stabilize training and avoid destructive policy shifts.
% To this end, we define the token-level loss by combining the normalized advantage with a KL divergence penalty. Specifically, for a token $o_{i,t}^j$, the loss is formulated as Eq.~\ref{eq:lit}.
% \begin{equation}
% \begin{split}
%     l_{i,t}^j = &\frac{\pi_\theta (o_{i,t}^j \mid p, o_{i<t})}{[\pi_\theta (o_{i,t}^j \mid p, o_{i<t})]_\text{no\_grad}}A_{i,t}^j \\
%     &- \beta D_{KL}(\pi_\theta \mid \mid \pi_\text{ref})
%     \label{eq:lit}
% \end{split}
% \end{equation}
% Here, $[\cdot]_\text{no\_grad}$ indicates that the denominator is detached from the computation graph during backpropagation, preventing gradient leakage and stabilizing training. $\beta$ is a hyperparameter controlling the strength of the KL regularization. 
Then, the KL divergence between the current policy and the reference policy is added as a penalty regularization term, with a hyperparameter $\beta$ to control the strength. This helps to further constrain the update policy from deviating too far from the initial model, ensuring training stability.
Finally, the overall GRPO loss is computed by first averaging the token-level losses within each sequence, and then taking the mean across all sequences in the batch. Formally, the total loss $\mathcal L$ is defined as Eq.~\eqref{eq:GRPO}
\begin{equation}
    \mathcal L = -\frac{1}{k}\sum_{j=1}^k \frac{1}{|o_i^j|}\sum_{t=1}^{|o_i^j|}\left(l_{i,t}^j - \beta D_\text{KL}(\pi_\theta\mid\mid \pi_\text{ref})_{i,t}^j\right)
    \label{eq:GRPO}
\end{equation}
After computing the final loss, we use it to update a LoRA adaptor, enabling efficient fine-tuning. {Details of the  prompts are provided in the App.~\ref{sec:pp}.}

%% file: table/eval.tex
\begin{tabular}{lcccc}
  \hline
  \textbf{Model} & \textbf{STA} & \textbf{SIM} & \textbf{FL} & \textbf{J} \\
  \hline
  Human reference & 95.53 & 77.33 & 88.23 & 65.36 \\
  \hline
  \multicolumn{5}{c}{\textit{Baseline Models}} \\
  \hline
  ParaDetox           & 90.31 & 85.77 & 88.97 & 67.83 \\
  DetoxLLM$^*$            & 95.38 & 59.15 & \textbf{97.62} & 54.70 \\
  YOPO                & 82.27 & \textbf{89.40} & 87.03 & 62.56 \\
  \hline
  T5 SFT              & 62.74 & \textit{88.64} & \textit{88.25} & 46.89 \\
  BART SFT            & 87.18 & 86.68 & 89.12 & 66.40 \\
  % LLaMA3.1-8B zero shot & 58.27 & \textit{52.44} & \textit{\textbf{98.98}} & 30.16 \\
  Llama3.1-8B SFT & 86.14 & 83.12 & 93.14 & 65.36 \\  
  Qwen2.5-7B SFT & 90.31 & 83.05 & 90.16 & 66.94 \\
  \hline
  % \ding{168}Qwen2.5-7B zero shot & \textbf{96.43}& 68.38 & 97.46 & 64.10 \\
  Llama3.1-8B five shot & 66.32 & \textit{58.58} & \textit{96.72} & 37.49 \\
  Qwen2.5-7B five shot & 89.12& 75.23 & 96.42 & 63.91 \\
  
  \hline
  \multicolumn{5}{c}{\textit{Our models}} \\
  \hline
  Llama3.1-8B+SFT+GRPO & \textbf{95.98} & 82.39 & 88.38 & \textbf{69.61} \\
  Qwen2.5-7B+SFT+GRPO & 93.74 & 83.93 & 87.33 & 68.26 \\
  \hline
\end{tabular}

%% file: table/ood.tex
\resizebox{0.85\textwidth}{!}{\begin{tabular}{lcccccccccc}
\hline
\textbf{Dataset} & \multicolumn{4}{c}{\textbf{DetoxLLM}} & \multicolumn{4}{c}{\textbf{HuggingFace}} \\
\textbf{Metric} & \textbf{STA} & \textbf{SIM} & \textbf{FL} & \textbf{J} & \textbf{STA} & \textbf{SIM} & \textbf{FL} & \textbf{J} \\
\hline
\textit{Original reference} & 97.13 & 44.93 & 98.17 & 42.97 & 60.20 & 74.02 & 89.40 & 40.00 \\
\hline
ParaDetox          & 59.27 & 84.20 & 87.21 & 41.43 & 78.00 & 91.92 & 93.20 & 65.50 \\
DetoxLLM$^*$           & \textbf{95.04} & 49.41 & \textbf{98.43} & 46.08 & \textbf{94.20} & 61.02 & \textbf{98.60} & 55.63 \\
YOPO               & 51.44 & \textbf{91.05} & 88.51 & 39.25 & 72.80 & \textbf{95.18} & 92.80 & 62.89 \\
\hline
T5 SFT           & 36.81 & \textit{86.50} & \textit{89.93} & 27.34 & 65.40 & \textit{91.41} & \textit{94.22} & 55.11 \\
BART SFT          & 59.53 & 81.78 & 89.56 & 39.62 & 77.40 & 89.94 & 94.40 & 64.61 \\
Llama3.1+SFT       & 64.49 & 78.46 & 91.38 & 44.47 & 86.40 & 83.88 & 96.40 & 68.73 \\
Qwen2.5+SFT        & 64.49 & 76.45 & 86.95 & 40.57 & 83.20 & 86.02 & 94.00 & 65.60 \\
\hline
Llama3.1+SFT+GRPO  & 71.02 & 83.29 & 88.25 & 51.34 & 89.00 & 85.43 & 96.20 & \textbf{72.65} \\
Qwen2.5+SFT+GRPO   & 73.89 & 79.57 & 90.34 & \textbf{51.94} & 91.20 & 84.09 & 94.00 & 71.39 \\
\hline
\end{tabular}}

%% file: table/ablation.tex
\resizebox{0.85\linewidth}{!}{\begin{tabular}{lcccc}
  \hline
  \textbf{Method} & \textbf{STA} & \textbf{SIM} & \textbf{FL} & \textbf{J} \\
  \hline
  Zero Shot                  & 58.27 & \textit{52.44} & \textit{\textbf{98.98}} & 30.16 \\
  % Five Shot & \textit{66.32} & \textit{58.58} & \textit{96.72} & \textit{54.93} \\
  GRPO                       & 78.09 & \textit{70.24} & \textit{96.86} & 52.73 \\
  SFT+GRPO                   & 94.78 & 82.32 & 87.18 & 67.61 \\
  Data Select+SFT+GRPO       & \textbf{95.98} & \textbf{82.39} & 88.38 & \textbf{69.61} \\
  \hline
\end{tabular}}

%% file: table/dataproportion.tex
\begin{table}[tbp]
\centering
\caption{Sensitivity analysis of model performance with varying proportions of training data during cold-start.}
\label{tab:data_proportion}
\resizebox{0.85\columnwidth}{!}{\begin{tabular}{c *{4}{S[table-format=2.2]}}
\toprule
{Data Proportion} & {STA} & {SIM} & {FL} & {J} \\
\midrule
10\% & 95.08 & 82.27 & 88.23 & 68.54 \\
20\% & 95.98 & 82.39 & 88.38 & 69.61 \\
30\% & 94.78 & 82.60 & 86.59 & 67.29 \\
40\% & 95.23 & 82.20 & 86.74 & 67.33 \\
\bottomrule
\end{tabular}}
\end{table}

% Table~\ref{tab:data_proportion} presents the performance of models trained with varying proportions of data during the cold-start phase. The experiments are conducted based on the LLaMA3.1-8B-Instruct model. All other hyperparameters and training configurations are kept consistent with those used in the final model. 
% The results indicate that despite minor fluctuations, overall performance remains stable across settings, with all configurations achieving a strong J-score (>67). This suggests that the proportion of cold-start data has limited influence on the final model’s effectiveness.
% Among the tested data proportions, 20\% not only yields the best performance but also strikes an optimal balance: it provides sufficient data for the Cold Start phase while avoiding excessive exposure during the SFT stage, which could otherwise undermine the effectiveness of subsequent GRPO training.
% Sensitivity analyses of other hyperparameters are provided in the Appendix~\ref{sec:PSA}.

To investigate the impact of the data selection ratio during the cold-start stage, we conduct a parameter sensitivity study based on the Llama3.1-8B-Instruct model. As illustrated in Tab.~\ref{tab:data_proportion}, results show that our method performs robustly across a range of cold-start data proportions between 10\% and 40\%, with the optimal performance achieved at 20\%.
When the data ratio falls below 20\%, the limited amount of initial data leads to a weaker foundation, resulting in a lower performance ceiling compared to the 20\% setting. 
Conversely, when using more than 20\% of the data, the model performance deteriorates despite acquiring stronger task-specific knowledge. This may be attributed to the model memorizing more correct responses during the SFT stage. As highlighted in DAPO~\cite{dapo}, if the model generates either entirely correct or entirely incorrect outputs for all samples in a group, the within-group advantage becomes zero. Consequently, such samples fail to contribute to parameter updates during GRPO, reducing the number of effective training signals and ultimately degrading performance.
% Another interesting observation is that when the data proportion exceeds 20\%, the fluency of generated sentences drops significantly. This may be attributed to the inherent incompleteness of sentences commonly found in online forums. With increased data used for supervised fine-tuning, the model may overfit to such patterns, leading to degraded fluency.
Sensitivity analyses of other hyperparameters including $\lambda$ and $\alpha$ are provided in the App.~\ref{sec:PSA}.

%% file: table/atest.tex
\begin{table}[h]
\centering
\caption{Performance of the model with different values of $\lambda$.}
\label{tab:lambda}
\begin{tabular}{ccccc}
\toprule
$\lambda$ & STA    & SIM    & FL     & J      \\
\midrule
1 & 92.85 & 84.54 & 85.54 & 67.23 \\
3 & 94.34 & 83.99 & 85.54 & 67.52 \\
5 & 95.98 & 82.39 & 88.38 & 69.61 \\
7 & 95.23 & 82.62 & 85.84 & 67.26 \\
\bottomrule
\end{tabular}
\end{table}

Table~\ref{tab:lambda} presents the performance of our model under different values of the weighting parameter $\lambda$ in the reward function, which balances detoxification effectiveness and semantic preservation. To ensure a fair comparison, all other settings were kept consistent with those of the final model.

From the experiments, we observe the following: 

(1)Our method is robust to variations in $\lambda$, as the joint score (J) consistently remains above 67 across all settings. This indicates stable performance regardless of the reward weighting.

(2)Increasing $\lambda$ generally leads to higher STA scores, indicating that the model is being successfully guided to prioritize detoxification. However, this comes at the cost of reduced SIM scores, as greater emphasis on detoxification in the reward function naturally leads to diminished semantic preservation---a trade-off that is expected in such multi-objective optimization settings.

(3)When $\lambda$ exceeds 5, the STA score no longer increases and instead slightly declines. A possible explanation for this is the discrepancy between the toxicity classifiers used during reward computation and final evaluation. Specifically, the toxicity classifier embedded in the reward function may be more lenient and fail to detect certain subtle toxicities that the STA evaluation classifier can identify. As a result, further increasing the detoxification weight in the reward has diminishing returns---since the reward function itself cannot distinguish those remaining toxic cases, the model is not penalized for them during training.

%% file: table/alpha.tex
\begin{table}[h]
\centering
\caption{Performance of the model with different values of $\alpha$.}
\label{tab:alpha}
\begin{tabular}{ccccc}
\toprule
$\alpha$ & STA    & SIM    & FL     & J      \\
\midrule
0.4 & 94.63 & 81.79 & 83.76 & 64.53 \\
0.5 & 95.98 & 82.39 & 88.38 & 69.61 \\
0.6 & 93.59 & 82.88 & 86.69 & 67.05 \\
\bottomrule
\end{tabular}
\end{table}

Table~\ref{tab:alpha} reports model performance across different values of $\alpha$, the filtering threshold defined in Equation~\ref{eq:1}. Values from 0.4 to 0.6 were tested, with all other settings identical to those of the final model. The 20\% subset of the dataset contains a total of 2,186 samples. After applying the filtering thresholds with $\alpha$ values of 0.4, 0.5, and 0.6, the remaining numbers of samples are 2,094, 2,011, and 1,869, respectively. 
The experimental results show that the overall model performance drops significantly when $\alpha = 0.4$, indicating that in the cold-start phase, the limited amount of data amplifies the negative impact of noisy samples, leading the model to learn spurious patterns. When $\alpha \geq 0.5$, model performance improves notably, suggesting that moderate filtering effectively removes low-quality data, enhances the purity of the training signal, and reduces the influence of noise. However, when $\alpha = 0.6$, certain challenging toxic samples may be disproportionately filtered out, leading to a distributional shift in the training data. This shift can hinder the model’s ability to generalize detoxification patterns, resulting in suboptimal performance. Overall, $\alpha = 0.5$ effectively filters out low-quality samples without disturbing the data distribution, leading to optimal detoxification performance.